\documentclass{article}

\PassOptionsToPackage{numbers, compress}{natbib}
\usepackage[preprint]{neurips_2025}

\usepackage{algorithm}
\usepackage{algorithmic}
\usepackage[utf8]{inputenc} % allow utf-8 input
\usepackage[T1]{fontenc}    % use 8-bit T1 fonts
\usepackage{hyperref}       % hyperlinks
\usepackage{url}            % simple URL typesetting
\usepackage{booktabs}       % professional-quality tables
\usepackage{amsfonts}       % blackboard math symbols
\usepackage{amsmath}
\usepackage{nicefrac}       % compact symbols for 1/2, etc.
\usepackage{microtype}      % microtypography
\usepackage{xcolor}         % colors
\usepackage{bm}
\usepackage[capitalize]{cleveref}
\usepackage{graphicx}
\usepackage{wrapfig}

\title{Semi-Implicit Variational Inference via Kernelized Path Gradient Descent}

\author{%
  Tobias Pielok, Bernd Bischl, David R{\"u}gamer \\
  Department of Statistics, LMU Munich, Munich, Germany\\
  Munich Center for Machine Learning, Munich, Germany\\
  \small\texttt{\{tobias.pielok, bernd.bischl, david.ruegamer\}@stat.uni-muenchen.de}\\
}

\newcommand{\R}{\mathbb{R}}
\newcommand{\N}{\mathbb{N}}

\newcommand{\E}{\mathbb{E}}
\newcommand{\V}{\mathbb{V}}

\newcommand{\KL}{D_{\mathrm{KL}}}

\newcommand{\supp}{\mathrm{supp}\,}

\newcommand{\z}{\bm{z}}

\newcommand{\zetatildev}{\widetilde{\bm{\zeta}}}
\newcommand{\zp}{\z^\prime}
\newcommand{\yv}{\bm{y}}

\newcommand{\eps}{\bm{\epsilon}}
\newcommand{\etav}{\bm{\eta}}
\newcommand{\zgiveneps}{\z\vert\eps}
\newcommand{\zgivenyv}{\z\vert\yv}
\newcommand{\epsgivenz}{\eps\vert\z}

\newcommand{\qz}{q_{\z}}

\newcommand{\pz}{p_{\z}}
\newcommand{\peps}{p_{\eps}}

\newcommand{\petav}{p_{\etav}}

\newcommand{\qzgiveneps}{q_{\zgiveneps}}
\newcommand{\qepsgivenz}{q_{\epsgivenz}}
\newcommand{\qzgivenyv}{q_{\zgivenyv}}
\newcommand{\pepsetav}{p_{\eps, \etav}}
\newcommand{\qzeps}{q_{\z, \eps}}

\newcommand{\phiv}{\bm{\phi}}
\newcommand{\fphi}{f_{\phiv}}
\newcommand{\hphi}{h_{\phiv}}

\newcommand{\thetav}{\bm{\theta}}

\newcommand{\ztilde}{\widetilde{\bm{z}}}

\newcommand{\sis}{s_{\mathrm{IS}, k}}
\newcommand{\tauepsgivenz}{\tau_{\epsgivenz}}
\newcommand{\tautildeepsgivenz}{\widetilde{\tau}_{\epsgivenz}}

\newtheorem{theorem}{Theorem}[section]
\newtheorem{proposition}[theorem]{Proposition}

\newtheorem{definition}[theorem]{Definition}

\begin{document}

\maketitle

\begin{abstract}
Semi-implicit variational inference (SIVI) is a powerful framework for approximating complex posterior distributions, but training with the Kullback–Leibler (KL) divergence can be challenging due to high variance and bias in high-dimensional settings. While current state-of-the-art semi-implicit variational inference methods, particularly Kernel Semi-Implicit Variational Inference (KSIVI), have been shown to work in high dimensions, training remains moderately expensive. In this work, we propose a kernelized KL divergence estimator that stabilizes training through nonparametric smoothing. To further reduce the bias, we introduce an importance sampling correction. We provide a theoretical connection to the amortized version of the Stein variational gradient descent, which estimates the score gradient via Stein's identity, showing that both methods minimize the same objective, but our semi-implicit approach achieves lower gradient variance. In addition, our method's bias in function space is benign, leading to more stable and efficient optimization. Empirical results demonstrate that our method outperforms or matches state-of-the-art SIVI methods in both performance and training efficiency.
\end{abstract}

\section{Introduction}
Accurately approximating complex probability distributions is fundamental for tasks such as Bayesian inference and learning in energy-based models, where distributions—such as Gibbs distributions—are defined in terms of an energy function. In such settings, latent variables often have physical or semantic meaning, and capturing the correct structure and uncertainty of the posterior is critical for robust learning and decision-making.

Variational Inference (VI) is a powerful framework for approximating complex posterior distributions in probabilistic models. Traditional or explicit VI relies on simple, tractable families of distributions and typically minimizes the Kullback-Leibler (KL) divergence. While computationally efficient, this approach can lead to biased approximations when the variational family is too restrictive. In contrast, implicit VI leverages flexible distributions defined by sampling procedures without requiring a tractable density, enabling more expressive posteriors but often relying on adversarial or score-based techniques.

Semi-implicit variational inference (SIVI) strikes the balance between expressivity and tractability by defining variational distributions as mixtures with an implicit component. 

Among the different SIVI methods (see \cref{sec:lit} for related literature), \cite{cheng2024kernel} proposed a score-based method called KSIVI, which achieves state-of-the-art performance using kernelized Stein discrepancies to estimate gradients of implicit variational distributions. In contrast to most previous SIVI methods, KSIVI can be efficiently trained and scales well to high-dimensional problems, establishing it as a practical and state-of-the-art approach.

\begin{wrapfigure}{r}{0.45\textwidth}
    \centering
    \includegraphics[width=0.9\linewidth]{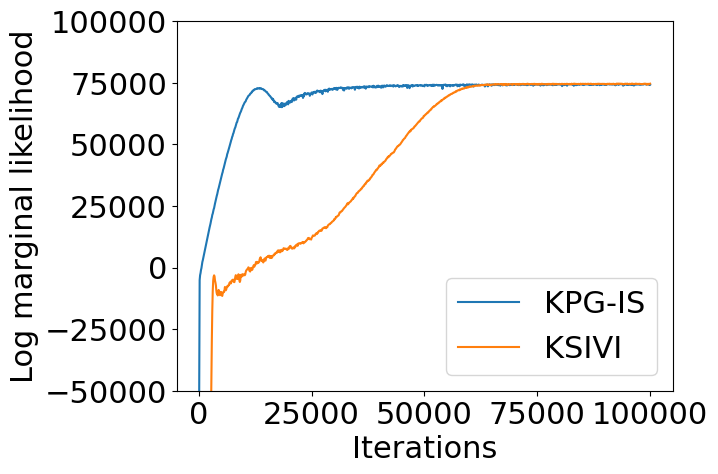}
    \caption{Convergence speed comparison between the current state-of-the-art method KSIVI and our proposal KPG-IS.}
    \label{fig:one}
\end{wrapfigure}

\paragraph{Our contributions} In this work, we propose a novel approach for performing inference with semi-implicit variational distributions by combining kernelized score estimation with pathwise gradients. Our contributions are as follows:
(i) We introduce the \emph{Kernelized Path Gradient (KPG)}, a method that leverages the reparameterization structure of semi-implicit distributions and enables efficient gradient-based optimization.
(ii) We show that KPG yields a provably lower-variance estimator than existing Stein-based methods.
(iii) To further improve sample efficiency and reduce bias, we introduce \emph{KPG-IS}, an importance-weighted variant that learns a proposal distribution for the latent variables via a constrained mixture model.
(iv) We show that the optimal proposal distribution involves a tradeoff between bias and variance, and can be learned for each specific tradeoff in an unbiased manner.
(v) Finally, we demonstrate empirical on-par performance or improvements over state-of-the-art semi-implicit inference methods (cf.~\cref{fig:one}).

\section{Background}

\subsection{Semi-implicit variational inference}
To generate a sample $\z$ from a semi-implicit distribution $\qz$ defined over $Z \subset \R^{d_Z}$ with $d_Z \in \N$, we first draw a latent random variable $\eps \sim \peps$ taking values in $E\subset\R^{d_E}$ with $d_E\in\N$, transform it through a neural network $\fphi:E\rightarrow Y$ with $Y\subset\R^{d_Y}$ and $d_Y\in\N$ to model complex dependencies, and use the output $\yv = \fphi(\eps)$ to parameterize a simple explicit distribution $\qzgivenyv$, such as a factorized Gaussian. Under the assumption that the conditional density is reparameterizable, there exists a function $g:Y\times H  \rightarrow Z$ such that a sample $\z$ can be generated by drawing a random variable $\etav$ taking values in $H \in \R^{d_H}$ with $d_H\in\N$ and then evaluating $g$ at $\etav$ and the parameters $\yv,$ i.e.,
\begin{equation}
\z = g(\yv, \etav) = \underbrace{g(\fphi(\eps), \etav)}_{=: \hphi(\eps, \etav)}.
\end{equation}
For example, in the case where the conditional distribution is a factorized Gaussian, sampling is performed by drawing $\etav \sim \mathcal{N}(0, I)$ and computing
\begin{equation}
\z = \bm{\mu}_{\eps} + \textrm{diag}(\bm{\sigma}_{\eps})  \etav,
\end{equation}
where $\bm{\mu}_{\eps}$ and $\bm{\sigma}_{\eps}$ denote the mean and standard deviation vectors, respectively, which in this case are the outputs of $\fphi$. 
This construction enables us to express the likelihood of a sample $\z$ in a principled manner, such that
\begin{equation}
    \qz(\z) = \E_{\eps\sim\peps}\left[\qzgivenyv\left(\z\vert\fphi(\eps)\right)\right] = \E_{\eps\sim\peps}\left[\qzgiveneps\left(\z\vert\eps\right)\right].
\end{equation}
Note also that expectations with respect to $\qz$ are compatible with the reparameterization trick \cite{Kingma2013AutoEncodingVB}; that is, for a differentiable function $\ell:Z\rightarrow\R$ which could possibly also depend on $\phiv$, it holds that 
\begin{equation}
    \nabla_{\phiv}\E_{\z \sim \qz}\left[\ell(\z) \right] = \E_{\eps, \etav \sim \pepsetav}\nabla_{\phiv}\left[\ell(\hphi(\eps, \etav)) \right].
\end{equation}

\subsection{Amortized Stein variational gradient descent}
Several amortized versions of \emph{Stein Variational Gradient Descent} (SVGD) have been introduced in \cite{FengWL17} to enable inference through neural networks. Here, we briefly describe the most widely used formulation, which views amortized SVGD as minimizing the kernel-smoothed difference between score functions. More specifically, suppose we have a neural sampler $\qz$, which means that we can generate samples by transforming noise $\bm{\xi} \sim p_{\bm{\xi}}$ through a neural network $\hphi$, but cannot evaluate the likelihood of the samples. The objective is to minimize the reverse KL divergence between $
\qz$ and the target distribution $\pz,$ i.e.,
\begin{equation}
    \KL(\qz \Vert \pz) = \E_{\z \sim \qz}\left[\log\left(\frac{\qz(\z)}{\pz(\z)}\right)\right].
\end{equation}
Since $\qz$ is amenable to the reparameterization trick, a low-variance gradient estimator of the KL divergence can be derived~\cite{Roeder2017Sticking}. Hence, leveraging the fact that the expected score function vanishes, we obtain the \emph{pathwise gradient estimator}, i.e.,

\begin{equation}
    \label{eq:pg}
        \nabla_{\phiv} \KL(\qz \Vert \pz) = \E_{\bm{\xi} \sim p_{\bm{\xi}}}\left[\Delta(\hphi(\bm{\xi}))  \cdot \nabla_{\phiv}\hphi(\bm{\xi})\right].
\end{equation} 
where the difference in score gradients $\Delta(\z) = \nabla_{\z}\log \qz(\z) - \nabla_{\z}\log \pz(\z).$ However, we do not have access to the score gradient $\nabla_{\z} \log \qz(\z).$ To address this, we first rewrite the score gradient difference such that
\begin{equation}
\label{eq:score_max}
    \Delta(\z) = \mathop{\mathrm{arg\;max}}_{\omega \in \mathcal{H}} \E_{\z\sim\qz}\left[2\omega
    (\z)^\top\Delta(\z) - \Vert \omega(\z)\Vert^2_{\mathcal{H}}\right],
\end{equation}
where the function space $\mathcal{H}$ is the space of square integrable functions $L^2.$
To apply Stein’s identity for estimating the score gradient \cite{li2018gradient,liusvgd2016}, we utilize the result proven in \cite{cheng2024kernel}, which states that when the function space $\mathcal{H}$ of the maximization problem defined in Eq.~\ref{eq:score_max} is restricted to a reproducing kernel Hilbert space (RKHS) with a given kernel function $k:Z\times Z\rightarrow \R_{\geq 0}$, the problem admits a unique solution, given explicitly by 
\begin{equation}
\label{eq:kpg}
    \Delta_k(\z) = \E_{\zp \sim \qz}k(\z, \zp)\left[\nabla_{\zp}\log \qz(\zp) - \nabla_{\zp}\log \pz(\zp)\right].
\end{equation}
Under the assumptions that the kernel $k$ is continuously differentiable and \begin{equation}
    k(\z,\zp)\qz(\zp)\vert_{\partial Z} = 0 \, \text{ or } \lim_{\zp\rightarrow \infty}k(\z,\zp)\qz(\zp) = 0 \, \text{ if } Z=\R^{d_Z},
\end{equation} we get that 
\begin{equation}
\label{eq:svgd}
    \Delta_{\textrm{STEIN},k}(\z) := \E_{\zp \sim \qz}\left[-\nabla_{\zp}k(\z, \zp) - k(\z, \zp)\nabla_{\zp}\log \pz(\zp)\right] = \Delta_k(\z)
\end{equation}
by using Stein's identity, which can be proved via integration by parts.
Replacing the difference in score gradients $\Delta$ in the 
 pathwise gradient estimator given by Eq.~\ref{eq:pg} with a Monte-Carlo estimate of the kernelized difference in score gradients $\Delta_k$ in Eq.~\ref{eq:svgd} results in the amortized SVGD update step.

\section{Method} \label{sec:method}

We note that the same derivation as amortized SVGD can be followed; however, rather than relying on a simple neural sampler, we leverage a semi-implicit distribution since we can just substitute $\bm{\xi}$ with $(\eps, \etav)$. While we cannot directly plug Eq.~\ref{eq:kpg} into Eq.~\ref{eq:pg}, we can first apply the kernel trick for semi-implicit distributions, as introduced by \cite{cheng2024kernel}, which states that 
\begin{equation}
    \E_{\zp \sim \qz}\left[k(\z, \zp)\nabla_{\zp}\log \qz(\zp)\right] = \E_{\zp, \eps^\prime \sim \qzeps}\left[k(\z, \zp)\nabla_{\zp}\log \qzgiveneps(\zp\vert\;\eps^\prime )\right].
\end{equation}
This enables us to compute a Monte Carlo estimate of the resulting expression, i.e.,
\begin{equation}
\label{eq:kpgsi}
    \Delta_{\textrm{SI},k}(\z) := \E_{\zp,\eps^\prime \sim \qzeps}k(\z, \zp)\left[\nabla_{\zp}\log \qzgiveneps(\zp\vert\;\eps^\prime ) - \nabla_{\zp}\log \pz(\zp)\right] \quad (= \Delta_k(\z)),
\end{equation}
which can then be substituted into Eq.~\ref{eq:pg}. With this, we define the Kernelized Path Gradient (KPG) as 
\begin{equation}
\label{eq:kpg_def}
    \E_{\eps,\etav \sim \pepsetav}\left[\Delta_{\textrm{SI},k}(\hphi(\eps, \etav))  \cdot \nabla_{\phiv}\hphi(\eps, \etav)\right].
\end{equation}
Exploiting the hierarchical structure of the semi-implicit distribution is a crucial distinction, as it reduces the variance of the score gradient estimator, avoids boundary assumptions, and requires only a continuous kernel.
To analyze the difference in variability between the Stein gradient estimator
\begin{equation}
s_{\textrm{STEIN},k}(\z) := \E_{\zp \sim \qz}\left[ -\nabla_{\zp}k(\z, \zp) \right]
\end{equation}
and the semi-implicit gradient estimator
\begin{equation}
s_{\textrm{SI},k}(\z) := \E_{\zp, \eps^\prime \sim \qzeps}\left[ k(\z, \zp)\nabla_{\zp}\log \qzgiveneps(\zp \mid \eps^\prime) \right],
\end{equation}
we consider the trace of the difference of the covariance matrices \begin{equation}
    \Delta\V := \mathrm{trace}\left(\V\left[\hat{s}_{\textrm{STEIN},k}(\z)\right] - \V\left[\hat{s}_{\textrm{SI},k}(\z)\right]\right)
\end{equation} of their corresponding Monte Carlo estimators $\hat{s}_{\textrm{STEIN},k}(\z)$ and $\hat{s}_{\textrm{SI},k}(\z)$, computed using $n$ i.i.d.\ samples.
Since both estimators share the same expectation, i.e.,
$\E\left[\hat{s}_{\textrm{STEIN},k}(\z)\right] = \E\left[\hat{s}_{\textrm{SI},k}(\z)\right],$
we show in Appendix~\ref{sec:proof_var} that
\begin{equation}
\Delta\V = \E\left\Vert\hat{s}_{\textrm{STEIN},k}(\z)\right\Vert^2_2 - \E\left\Vert\hat{s}_{\textrm{SI},k}(\z)\right\Vert^2_2
\end{equation}
and establish the following proposition.

\begin{proposition}
\label{prop:var_diff}
    Assuming that the kernel $k$ is the Gaussian density kernel, i.e., $k(\z,\zp) = \frac{1}{(2\pi\sigma_k^2)^{d_z/2}}\exp\left(-\frac{\left\Vert\z - \zp\right\Vert^2_2}{2\sigma_k^2}\right)$ and $\qzgiveneps$ is a conditional Gaussian distribution, it holds for $\z \in Z$ that
    \begin{align}
    \label{eq:var_diff}
    \begin{split}
    \E\left\Vert\hat{s}_{\textrm{STEIN},k}(\z)\right\Vert^2_2& - \E\left\Vert\hat{s}_{\textrm{SI},k}(\z)\right\Vert^2_2 = \frac{1}{n}\E_{\eps^\prime, \etav^\prime \sim \pepsetav}k(\z, \mathrm{diag}(\bm{\sigma}_{\eps^\prime}) \etav^\prime + \bm{\mu}_{\eps^\prime})^2 \\
    &\cdot \left[\frac{\Vert  \mathrm{diag}(\bm{\sigma}_{\eps^\prime}) \etav^\prime + \bm{\mu}_{\eps^\prime} - \z\Vert^2_2 }{\sigma_k^4} - \Vert \mathrm{diag}(\bm{\sigma}_{\eps^\prime})^{-1} \etav^\prime\Vert^2_2\right].
       \end{split}
    \end{align}
\end{proposition}

To better understand how the difference in the variability of the score gradients depends on $\sigma_k$ and $\bm{\sigma}_{\eps^\prime}$, we first prove in Appendix~\ref{sec:proof_low_var_suf} a sufficient condition under which $\E\left\Vert\hat{s}_{\textrm{STEIN},k}(\z)\right\Vert^2_2 \geq \E\left\Vert\hat{s}_{\textrm{SI},k}(\z)\right\Vert^2_2$, providing insight into the scaling behavior with respect to these parameters.

\begin{proposition}
\label{prop:var_exp}
    Under the assumptions of Proposition~\ref{prop:var_diff}, it holds that $\E\left\Vert\hat{s}_{\textrm{STEIN},k}(\z)\right\Vert^2_2 \geq \E\left\Vert\hat{s}_{\textrm{SI},k}(\z)\right\Vert^2_2$ if
    \begin{equation}
        \min(\bm{\sigma}_{\eps^\prime})^4 - 2\min(\bm{\sigma}_{\eps^\prime})^2\max(\bm{\sigma}_{\eps^\prime})\frac{\Vert  \bm{\mu}_{\eps^\prime} - \z\Vert_2}{\Vert\etav^\prime\Vert_2}  +  \min(\bm{\sigma}_{\eps^\prime})^2\frac{\Vert  \bm{\mu}_{\eps^\prime} - \z\Vert^2_2}{\Vert\etav^\prime\Vert^2_2}\geq \sigma_k^4 \quad \text{a.s.}
    \end{equation}
\end{proposition}
Assuming that the maximum value of the random vector $\bm{\sigma}_{\eps^\prime}$ is less than one a.s., we observe that $\sigma_k$ scales benignly with $\bm{\sigma}_{\eps^\prime}$. Specifically, for sufficiently small maximum values of $\bm{\sigma}_{\eps^\prime}$, the term involving $\min(\bm{\sigma}_{\eps^\prime})^2$ is expected to dominate the inequality since, in general, $\bm{\mu}_{\eps^\prime} \neq \z$, suggesting that $\sigma_k$ only needs to scale quadratically a.s. with $\min(\bm{\sigma}_{\eps^\prime})$. This enables us to find an upper bound for the difference in the variance of the score gradient norms, with the proof provided in Appendix~\ref{sec:proof_ub_diff_sgv}.

\begin{proposition}
 \label{prop:var_diff_ub}
     Under the assumptions of Proposition~\ref{prop:var_exp} and additionally assuming that $k(\z, \zp)^2$ and $\left[\frac{\Vert  \mathrm{diag}(\bm{\sigma}_{\eps^\prime}) \etav^\prime + \bm{\mu}_{\eps^\prime} - \z\Vert^2_2 }{\sigma_k^4} - \Vert \mathrm{diag}(\bm{\sigma}_{\eps^\prime})^{-1} \etav^\prime\Vert^2_2\right]$ are negatively correlated, it holds that
         \begin{align}
    \label{eq:var_diff_ub}
    \begin{split}
    \E\left\Vert\hat{s}_{\textrm{STEIN},k}(\z)\right\Vert^2_2& - \E\left\Vert\hat{s}_{\textrm{SI},k}(\z)\right\Vert^2_2 \leq \frac{1}{n}\E_{\eps^\prime, \etav^\prime \sim \pepsetav}k(\z, \mathrm{diag}(\bm{\sigma}_{\eps^\prime}) \etav^\prime + \bm{\mu}_{\eps^\prime})^2\\
    &\cdot\underbrace{\left(\frac{\E_{\eps^\prime \sim \peps}\left[\Vert  \bm{\sigma}_{\eps^\prime}\Vert^2_2 + \Vert\bm{\mu}_{\eps^\prime} - \z\Vert^2_2\right] }{\sigma_k^4} - \E_{\eps^\prime \sim \peps}\left[\Vert \bm{\sigma}_{\eps^\prime}^{\odot-1}\Vert^2_2\right]\right)}_{=: \gamma} \\
    &\approx \frac{\qz(\z)}{n(2\sqrt{\pi}\sigma_k)^d} \cdot \gamma
       \end{split}
    \end{align}
    where $(\cdot)^\odot$ denotes the element-wise power.
\end{proposition}
 The upper bound becomes tight—that is, the inequality turns into an equality—when the correlation is zero. The assumption that $k(\z, \zp)^2$ and $\left[\frac{\Vert  \mathrm{diag}(\bm{\sigma}_{\eps^\prime}) \etav^\prime + \bm{\mu}_{\eps^\prime} - \z\Vert^2_2 }{\sigma_k^4} - \Vert \mathrm{diag}(\bm{\sigma}_{\eps^\prime})^{-1} \etav^\prime\Vert^2_2\right]$ are negatively correlated is plausible, since the first term is monotonically decreasing in $\Vert\z - \zp\Vert^2_2$, while the later term is monotonically increasing in $\Vert\z - \zp\Vert^2_2$. As $\max(\bm{\sigma}_{\eps^\prime})$ decreases, the correlation between the two terms weakens, since the first term becomes asymptotically independent of $\etav$, while the second term becomes increasingly dominated by it.
 Therefore, we expect the upper bound to be quite sharp in practice. Finally, we arrive at the approximate upper bound by noting that $(2\sqrt{\pi}\sigma_k)^d\cdot k^2$ is also a normalized kernel, for which the corresponding expectation converges to $\qz(\z)$ when $\sigma_k$ is sufficiently small. From this, two key insights emerge: first, a small $\max(\bm{\sigma}_{\eps^\prime})$ necessitates a correspondingly small $\sigma_k$; second, we expect that $\sigma_k$ only needs to scale quadratically with $\max(\bm{\sigma}_{\eps^\prime})$. As a result, the inequality becomes less restrictive due to the partial control we have over $\sigma_k$.
 
 Also, note this flexibility is beneficial because a small $\sigma_k$ is desirable—it leads to a more expressive RKHS $\mathcal{H}$, thereby reducing the bias introduced by the restriction to that RKHS.
Although this might suggest using a minimal kernel width, doing so naively results in a high-variance estimator, as only a few nearby samples contribute significantly to the estimate. Moreover, in high-dimensional settings, the curse of dimensionality further exacerbates this issue, since the number of samples required to populate a local neighborhood adequately grows exponentially with the dimension, making even a large number of samples potentially insufficient.

\subsection{Reducing the bias via importance sampling}
In light of these considerations, importance sampling offers a natural approach to mitigating variance. Yet, this method cannot be employed directly, since the likelihood $\qz$ of a semi-implicit distribution is intractable. However, note that we can write $\Delta_k$ such that
\begin{equation}
     \Delta_{\textrm{SI-IS},k}(\z) =  \E_{\eps\sim\tauepsgivenz}\E_{\zp \sim \qzgiveneps}\frac{\peps(\eps)k(\z, \zp)}{\tauepsgivenz(\epsgivenz)}\left[\nabla_{\z}\log \frac{\qzgiveneps(\zp\vert\eps)}{\pz(\zp)}\right] \quad (= \Delta_k(\z)),
\end{equation}
where $\tauepsgivenz$ is a conditional explicit distribution with $\supp \tauepsgivenz \supset \supp \peps$. This means that while the direct application of importance sampling is not feasible, it can still be applied to the latent variable $\eps$. This, in turn, raises the question of how to choose an optimal proposal distribution for the latent variable. Although there is no single correct choice, we adopt the following definition of optimality as it reflects a trade-off between reducing the bias by increasing density near the sample $  \z $—and controlling global importance sampling variance by limiting deviation from the latent distribution. The loss is naturally induced by the likelihood and remains fully differentiable.

\begin{definition}[Optimal proposal via mixture parameterization]
We call a proposal distribution $ \tauepsgivenz^* $ optimal, for $ \alpha(\z) \in (0, 1) $, if and only if it is parameterized as a convex combination of the latent distribution $ \peps $ and a learnable distribution $ \tautildeepsgivenz $, i.e.,
\begin{equation}
     \tauepsgivenz(\epsgivenz) = \alpha(\z) \, \peps(\eps) + (1 - \alpha(\z)) \, \tautildeepsgivenz(\epsgivenz),
\end{equation}
and minimizes the expected negative log-likelihood under the joint $ \qzeps $,
\begin{equation}
\label{eq:prop_objective}
    \tautildeepsgivenz^* \in \arg\min_{\tautildeepsgivenz} \,
\E_{\z, \eps \sim \qzeps} \left[
- \log \left(\tauepsgivenz(\epsgivenz)\qz(\z)\right)
\right].
\end{equation}
\end{definition}

This formulation encourages proposals that interpolate between being likely under the latent prior and being adapted to samples $ \z \sim \qz $.

We show in Appendix~\ref{sec:proof_rc1o} the following proposition.
\begin{proposition}
    The reverse conditional distribution $\qepsgivenz$ gives the strict lower bound of our objective, i.e., the optimal distribution when $\alpha(\z)$ converges to zero, and $\peps$ gives the trivial strict upper bound of our objective as $\alpha(\z)$ converges to one.
\end{proposition}
This highlights the motivation to use the hard constraint via mixture parametrization since only
\begin{equation}
    \supp \qepsgivenz = \supp \peps \underbrace{\frac{\qzgiveneps}{\qz}}_{\geq 0} \subset \supp \peps
\end{equation}
holds in general, while we require $\supp \tauepsgivenz \supset \supp \peps$ which is guaranteed for any $\alpha(\z) \in (0,1].$ Therefore, a small $\alpha(\z)$ is possible, but it increases the importance weight upper bound to \begin{equation}
   \lim_{\tautildeepsgivenz(\epsgivenz)\rightarrow 0}\; \frac{\peps(\eps)}{\alpha(\z)\peps(\eps) + (1-\alpha(\z))\tautildeepsgivenz(\epsgivenz)} = \frac{1}{\alpha(\z)} \leq \frac{1}{\underline{\alpha}},
\end{equation} where $\underline{\alpha} = \inf_{\z \in Z} \alpha(\z)$ likely impacting the bias-variance tradeoff. If $\tautildeepsgivenz$ is arbitrarily flexible, one optimal solution for $\alpha(\z)$ under our loss is always zero. However, when $\tautildeepsgivenz$ lacks sufficient expressiveness, choosing $\alpha(\z) > 0$ can result in a solution closer to the optimum. Therefore, we choose to learn $\alpha(\z)$ while enforcing the constraint \begin{equation}
    \alpha(\z) = \underline{\alpha} + (1-\underline{\alpha})\cdot\varsigma(\widetilde{\alpha}(\z)) \in (\underline{\alpha}, 1),
\end{equation} where $\underline{\alpha} \in (0,1)$, $\varsigma$ is the sigmoid function, and $\widetilde{\alpha}:Z\rightarrow\R$ is an unbounded function. This parametrization enables improved approximation while maintaining stability.

 Although the marginal $ \qz $ is inaccessible, the gradient of the objective given by Eq.~\ref{eq:prop_objective} with respect to the joined parameters $\thetav$ of $ \tautildeepsgivenz $ and $\widetilde{\alpha}$ can be expressed as
\begin{equation}
    - \E_{\z, \eps \sim \qzeps} \left[
 \nabla_{\thetav} \log \left( \alpha(\z) \, \peps(\eps) + (1 - \alpha(\z)) \, \tautildeepsgivenz(\epsgivenz) \right)
\right],
\end{equation}
which can be estimated via Monte Carlo without bias.

\subsection{Algorithms}
The algorithms developed in this work are introduced below. For both methods, the kernel width is determined in each iteration using the median heuristic \cite{liusvgd2016}.

\subsubsection{KPG}
 We derive a straightforward Monte Carlo estimator from the KPG given by Eq.~\ref{eq:kpg_def}. The corresponding procedure is detailed in Algorithm~\ref{alg:kpg}. This baseline is used to ablate the effect of importance sampling, which is a key component of our main method.

\begin{algorithm}[tb]
   \caption{KPG}
   \label{alg:kpg}
\begin{algorithmic}
   \STATE {\bfseries Input:}  target density $\pz$, kernel function $k,$ batch size $m,$ SIVI model $\hphi$
   \STATE $i=1,\dots,m,\quad l=1,2$
   \REPEAT
   \STATE $\eps_{i, l} \sim \peps, \etav_{i, l} \sim \petav$
   \STATE $\z_{i, l} = \hphi(\eps_{i, l}, \etav_{i, l})$
   \STATE $\ztilde_{i, l} = \texttt{stop\_gradient}(\z_{i, l})$
   \STATE $\widetilde{\Delta}_i(\ztilde_{i, 2}) = \nabla_{\ztilde_{i, 2}}\log \qzgiveneps(\ztilde_{i, 2}\vert\eps_{i, 2}) - \nabla_{\ztilde_{i, 2}}\log \log\pz(\ztilde_{i, 2})$ 
    \STATE $\textrm{loss} = 1/m^2 \sum^m_{j=1}(\sum^m_{i=1} k(\ztilde_{j, 1}, \ztilde_{i, 2}) \cdot \widetilde{\Delta}_i(\ztilde_{i, 2}))^\top \z_{j, 1}$
    \STATE $\phiv = \texttt{opt}(\mathrm{loss}, \phiv)$
   \UNTIL{$\phiv$ has converged}
\end{algorithmic}
\end{algorithm}

\subsubsection{KPG-IS}
Furthermore, we propose KPG-IS, a variant of the kernel path gradient (KPG) method enhanced via importance sampling. KPG-IS alternates between minimizing the expected forward KL divergence, $E_{\z\sim\qz}\left[\KL(\qepsgivenz\Vert\tauepsgivenz)\right]$, and the kernelized reverse KL divergence by following the KPG defined in Eq.~\ref{eq:kpg_def}. To estimate the kernelized score gradient difference  $\Delta_k(\z)$, we use the importance-weighted estimator $\Delta_{\textrm{SI-IS},k}(\z)$, which treats $\tauepsgivenz$ as the proposal distribution. This alternating optimization is enabled by the fact that $\sis(\z)$ is a consistent estimator of the score gradient whenever $\mathrm{supp}(\peps) \supset \mathrm{supp}(\tauepsgivenz)$. This support condition is guaranteed by our mixture parametrization, ensuring that the estimator remains valid throughout the training.

\begin{algorithm}[h]
   \caption{KPG-IS}
   \label{alg:kpg_is}
\begin{algorithmic}
   \STATE {\bfseries Input:}  target density $\pz,$ kernel function $k,$ batch size $m,$ number of latent samples $l,$ SIVI model $\hphi,$ conditional latent model $\tauepsgivenz,$ mixture coefficient NN $\alpha$
    \STATE $i=1,\dots,m, \quad j=1,\dots,l$
   \REPEAT
   \STATE $\eps_i \sim \peps, \etav_i \sim \petav$
   \STATE $\z_i = \hphi(\eps_i, \etav_i)$
   \STATE $\textrm{loss}_{\textrm{proposal}} = -1/m\sum^m_{i=1} \log \tauepsgivenz(\eps_i\vert\z_i, \alpha(\z_i))$
   \STATE $\thetav = \texttt{opt}(\textrm{loss}_{\textrm{proposal}}, \thetav)$
   \STATE
   \STATE $\eps_{i, j} \sim \tauepsgivenz(\cdot\vert\z_i), \etav_{i, j} \sim \petav$
   \STATE $\zetatildev_{i,j} = \texttt{stop\_gradient}(\hphi(\eps_{i, j}, \etav_{i,j}))$
   \STATE $\ztilde_{i} = \texttt{stop\_gradient}(\z_i)$
   \STATE $\log \widetilde{w}_{i, j} = \log k(\ztilde_i, \zetatildev_{i,j}) + \log \peps(\eps_{i, j}) - \log \tauepsgivenz(\eps_{i,j}\vert \ztilde_i, \alpha(\ztilde_i))$ 
   \STATE $\widetilde{\Delta}_{i,j}(\zetatildev_{i, j}) = \nabla_{\zetatildev_{i, j}}\log \qzgiveneps(\zetatildev_{i, j}\vert\eps_{i, j}) - \nabla_{\zetatildev_{i, j}}\log\pz(\zetatildev_{i, j})$ 
    \STATE $\textrm{loss} = 1/(m\cdot l) \sum^m_{i=1}(\sum^l_{j=1} \exp(\log \widetilde{w}_{i, j}) \cdot \widetilde{\Delta}_i(\zetatildev_{i, j}))^\top \z_{i}$
    \STATE $\phiv = \texttt{opt}(\mathrm{loss}, \phiv)$
   \UNTIL{$\phiv$ has converged}
\end{algorithmic}
\end{algorithm}

\section{Related literature} \label{sec:lit}

\cite{pmlr-v80-yin18b} introduced semi-implicit variational inference (SIVI), training models by sandwiching the ELBO between upper and lower bounds. \cite{pmlr-v89-titsias19a} later proposed a related ELBO-based objective with an unbiased gradient estimator, though it requires computationally intensive MCMC sampling. \cite{Sobolev2019Importance} advanced this direction by incorporating importance sampling into the SIVI framework.

Amortized Stein Variational Gradient Descent \cite{FengWL17} minimizes the same objective as semi-implicit methods but uses the Stein identity to compute the score gradient term. However, it does not explicitly leverage the semi-implicit structure, which can lead to higher variance in the gradient estimates due to the absence of such structural constraints.

\cite{lim2024particle} introduced Particle Semi-Implicit Variational Inference (PVI), which approximates Euclidean-Wasserstein gradient flows using a particle-based approach and has shown promising empirical results. Meanwhile, \cite{yu2023semiimplicit} proposed an alternative to ELBO-based training by minimizing the Fisher divergence. However, their minimax formulation introduces significant optimization challenges.

Building on this idea, \cite{cheng2024kernel} replaced the Fisher divergence with the kernel Stein discrepancy, transforming the minimax objective into a standard minimization problem. We refer to this method as KSIVI, which currently stands out as the gold standard among semi-implicit variational inference methods, achieving state-of-the-art performance while maintaining computational efficiency.

\section{Experiments} \label{sec:experiments}

We now turn to the empirical evaluation of our method. 
Following recent work in SIVI, the first problem is a common benchmark to test efficacy in high dimensions based on a diffusion process and initially analyzed in \cite{cheng2024kernel}. Our second experiment tackles a Bayesian linear regression model proposed by  \cite{pmlr-v80-yin18b}. Further common SIVI benchmarks can be found in the Appendix~\ref{app:furtherbenchmarks}. For all experiments involving KPG-IS, the proposal model $\tauepsgivenz$ is modeled as a Gaussian with a diagonal covariance structure, where the conditional parameters are learned using a neural network. As comparison methods, we use PVI \cite{lim2024particle}, as well as KSIVI \cite{cheng2024kernel}, which together form the current state-of-the-art in SIVI methods. For a fair comparison, all SIVI methods use the same neural network architecture, details of which are provided in the Appendix~\ref{app:expdetails}. We implemented KPG and KPG-IS
in PyTorch \cite{NEURIPS2019_9015}. 
All experiments are performed on a Linux-based server A5000 server with 2 GPUs, 24GB VRAM, and Intel Xeon Gold 5315Y processor with 3.20 GHz.

\subsection{Conditional diffusion process}

\begin{table}[h]
\caption{Log marginal likelihood estimates on the conditional diffusion benchmark. The reference estimate is computed using 1000 high-quality SGLD samples, while each method’s estimate is based on 60,000 samples.}
\label{tab:mllik_cond}
\begin{center}
\begin{sc}
\begin{small}
\begin{tabular}{lccc}
\toprule
Method & $\uparrow$ Log ML  & $\downarrow$ Seconds per iteration & Iterations  \\
\midrule
KPG-IS & $\mathbf{74528}$ & $1.45 \times 10^{-2}$ & $100$k\\
KPG    & $74311$ &  $2.50 \times 10^{-3}$ & $100$k\\
KSIVI  & $74504$ & $1.40 \times 10^{-2}$ & $100$k\\ 
STEIN &  $70371$ & $2.50 \times 10^{-3}$  & $100$k\\
PVI  & $47853$ &  $5.00 \times 10^{-3}$ & $100$k\\

\bottomrule
\end{tabular}
\end{small}
\end{sc}

\end{center}
\end{table}

To evaluate performance in high-dimensional settings, we consider a conditional diffusion process benchmark introduced in \cite{cheng2024kernel}, which has been used to assess the effectiveness of SIVI methods. It is based on the stochastic differential equation 
\begin{equation}
    dx_t = 10x_t(1 - x_t^2)dt + dw_t, \quad 0 \leq t \leq 1,
\end{equation}
with $ x_0 = 0 $ and $ w_t $ a standard Brownian motion \cite{10.5555/3327546.3327591}. After discretization of the SDE using the Euler-Maruyama scheme, one obtained a 100-dimensional latent variable $\bm{x}$ with prior $ p(\bm{x}) $ and observations $\bm{y}$ obtained by perturbing 20 time points of $\bm{x}$ using Gaussian noise. Accordingly, the likelihood $p(\bm{y}|\bm{x})$ is based on a Gaussian distribution assumption with 
\begin{equation}
    p(y_i|x_i) = (2\pi\sigma^2)^{-0.5} \exp((2\sigma^2)^{-1}(y_i-x_i)^2) \text{ with } \sigma^2 = 0.1.
\end{equation}
 The goal is to approximate the posterior $p(\bm{x}|\bm{y})$. As before, the ground truth is generated using SGLD with $100{,}000$ iterations and 1000 independent particles. The step size is again chosen to $0.0001$.
We again use the settings suggested for comparison methods as discussed in the literature \cite{cheng2024kernel}. In addition to PVI and KSIVI, we also run the amortized SVGD approach (STEIN) by \cite{FengWL17}.
%}

\begin{figure}[t]
    \centering
    \includegraphics[width=0.99\linewidth]{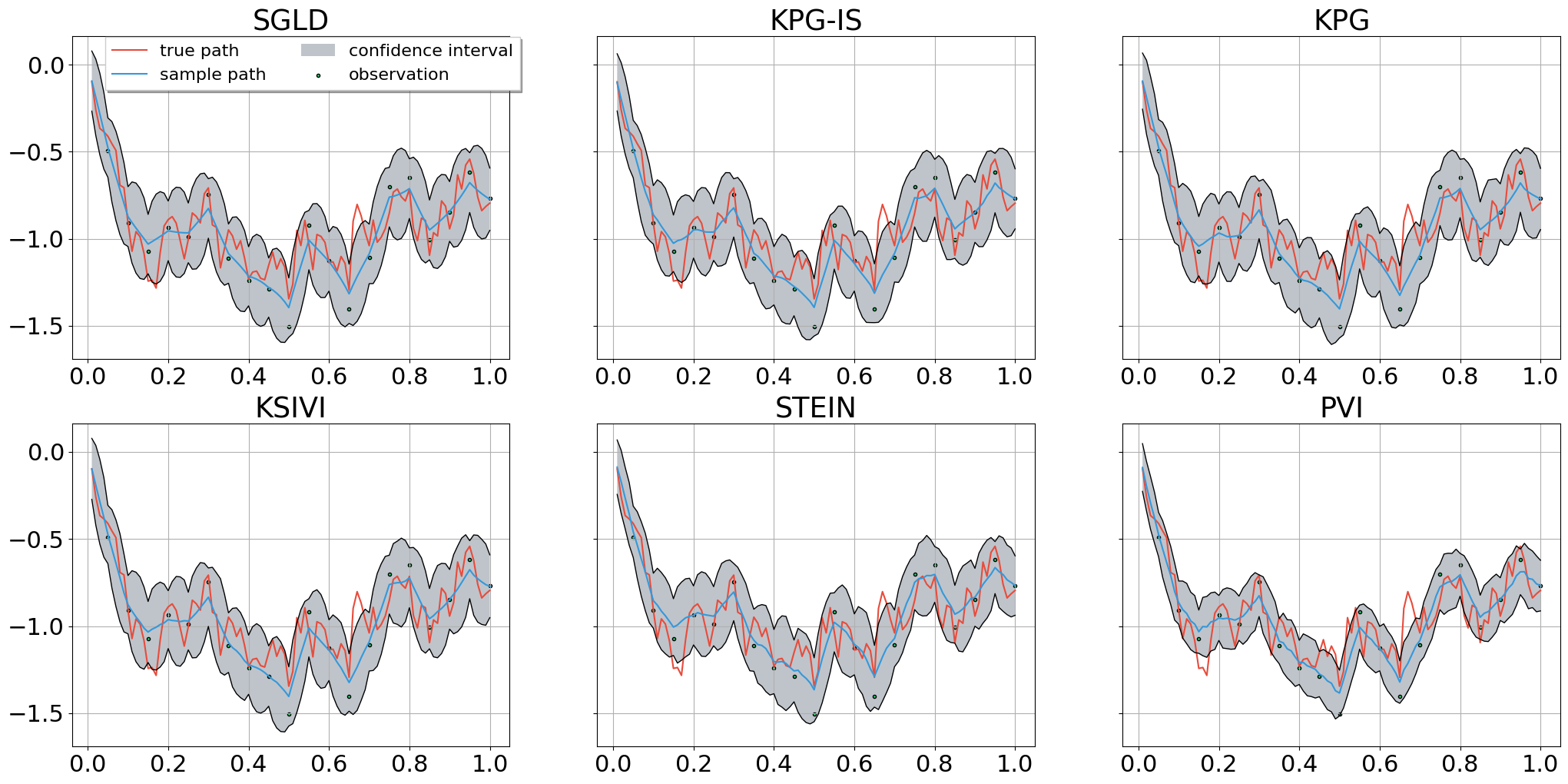}
    \caption{Comparison of posterior quality of different models (facets) for the diffusion process.}
    \label{fig:diff}
\end{figure}

\paragraph{Results} \cref{fig:diff} summarizes the results by showing the sample path and estimated confidence interval of each method. Using SGLD as a ground truth, we see that most methods perform well. However, notably less variation in the posterior is observed for PVI and also for STEIN. KPG-IS, KPG, and KSIVI perform similarly well, demonstrating that our method is on par with the current state-of-the-art. \cref{fig:one} exemplarily depicts one characteristic runtime comparison between KSIVI and KPG-IS. While both methods reach the same value, KPG is notably faster in convergence (cf.~\cref{fig:one} and \cref{tab:mllik_cond}). Further replications of this phenomenon can be found in the Appendix~\ref{app:cdp}.

\subsection{Bayesian linear regression}

\begin{figure}[h]
    \centering
    \includegraphics[width=1.0\linewidth]{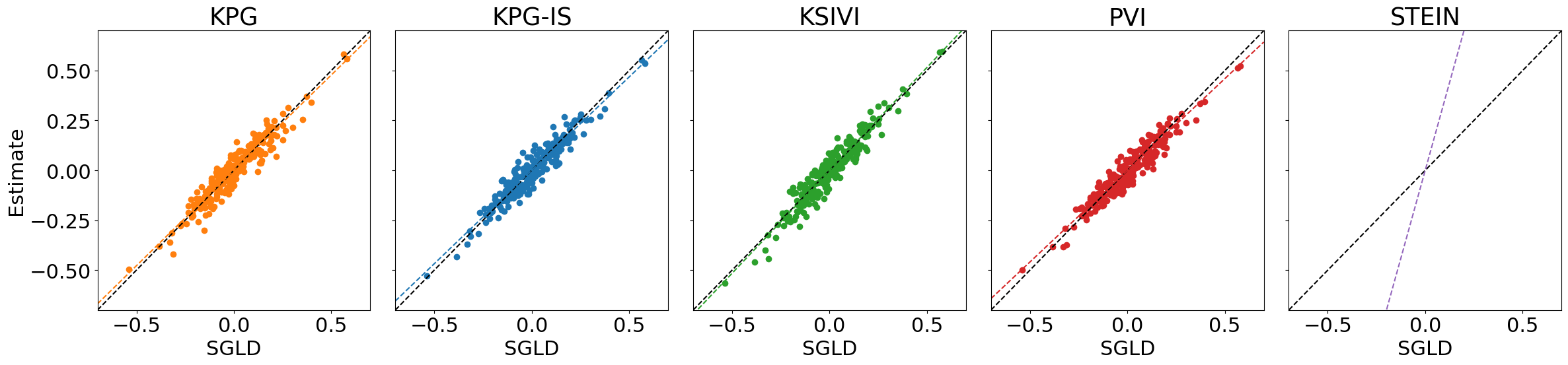}
    \caption{A scatter plot of all pairwise correlation coefficients $\rho_{i,j}$ between our estimates and those obtained from SGLD. The identity line indicates perfect agreement.}
    \label{fig:lr-cor}
\end{figure}

Another commonly used benchmark experiment for SIVI methods in moderate dimensions is a Bayesian logistic regression on the WAVEFORM dataset, which can be obtained from the UCI repository \cite{Dua:2019}. This problem was initially proposed in \cite{pmlr-v80-yin18b}. Given $y_i \in \{0, 1\}, i=1,\dots, N$ with $N=400$ and features $\bm{x}_i \in \R^{21},$ the model is defined by likelihood
\begin{align*}
       \ell(\bm{\beta}) = p(\bm{y}\vert \bm{X}; \bm{\beta}) = \prod^N_{i=1} \exp(\bm{x}_i^\top\bm{\beta})^{y_i} (1- \exp(\bm{x}_i^\top\bm{\beta}))^{1-y_i},
\end{align*}
where the goal is inference for the latent variable $\bm{\beta} \in \R^{22}$. As prior $p(\bm{\beta})$ an uninformative normal distribution with mean zero and variance 100 is used. The ground truth for this example is obtained by simulating stochastic gradient Langevin dynamics (SGLD) \cite{Welling2011BayesianLV}. Following \cite{cheng2024kernel}, we use $400{,}000$ iterations and 1000 samples for SGLD with a step size of 0.0001. We use the setup and optimized hyperparameters as in \cite{cheng2024kernel} to ensure a fair comparison. 

\textbf{Results}\,

\begin{figure}[h]
    \centering
    \includegraphics[width=0.9\linewidth]{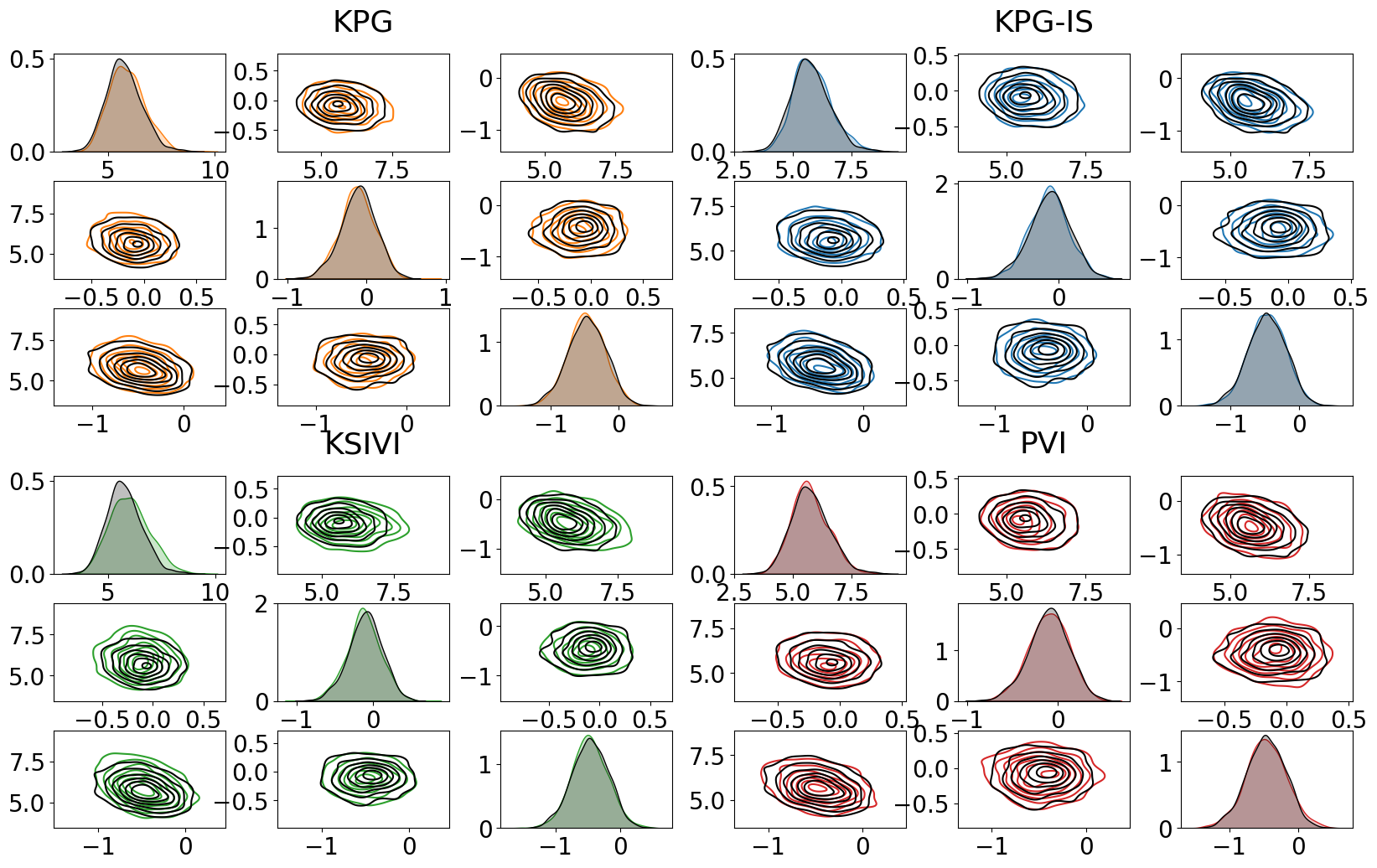}
    \caption{Comparison of marginal and pairwise density estimates for $\bm{\beta}^{(1)}$, $\bm{\beta}^{(2)}$, and $\bm{\beta}^{(3)}$, with SGLD estimates shown in black for reference.}
    \label{fig:lr-coef}
\end{figure}

Figures \ref{fig:lr-cor} and \ref{fig:lr-coef} show that all SIVI methods perform similarly in this example, with the exception of amortized SVGD, which diverges under the exact same hyperparameters used for KPG. This divergence, shown in the Appendix~\ref{app:blr}, highlights the impact of higher variance in the score gradient estimates. No systematic over- or underestimation of variances and correlations can be observed among the remaining methods. For KPG-IS, we set $\underline{\alpha} = 0.99$ and reused the uninformed $\eps$-samples, as detailed in Appendix\ref{app:impdetails}, to maintain computational efficiency, as target density evaluations are significantly more expensive in this example compared to the previous one.

\section{Discussion} \label{sec:discussion}
Our results demonstrate that semi-implicit variational inference with kernelized path gradients consistently outperforms amortized Stein variational methods across benchmarks, highlighting the benefits of explicitly leveraging the semi-implicit structure. The variance reduction achieved by KPG translates into more stable optimization and more accurate posterior approximations, particularly in challenging high-dimensional settings. Compared to the previous state-of-the-art method, KSIVI, our approach achieves comparable performance with significantly improved computational efficiency, making it a practical and scalable choice for complex models.
\subsection*{Limitations} 
The proposed method does not inherently promote exploration, which may limit its effectiveness in capturing complex distributions. While this is a common limitation shared with many related approaches, it remains an important area for improvement. In principle, our method can be combined with techniques such as temperature annealing \cite{pmlr-v37-rezende15} to encourage better mode coverage, but a more principled and integrated exploration mechanism would be desirable. Addressing this limitation is a promising direction for future research.

\newpage
\small

\bibliography{main}

\newpage

\appendix

\section{Proofs} \label{app:proofs}
\subsection{Variance comparison of score gradient estimators}
\label{sec:proof_var}
First note since $\E\left[\hat{s}_{\textrm{STEIN},k}(\z)\right] = \E\left[\hat{s}_{\textrm{SI},k}(\z)\right]$ it follows that 
\begin{align}
  \Delta\V :=  \mathrm{trace}&\left(\V\left[\hat{s}_{\textrm{STEIN},k}(\z)\right] - \V\left[\hat{s}_{\textrm{SI},k}(\z)\right]\right) \\
   &=\mathrm{trace}\left(\E\left[\hat{s}_{\textrm{STEIN},k}(\z)\hat{s}_{\textrm{STEIN},k}(\z)^\top\right] - \E\left[\hat{s}_{\textrm{SI},k}(\z)\hat{s}_{\textrm{SI},k}(\z)^\top\right]\right)\\
   &= \E\left[\hat{s}_{\textrm{STEIN},k}(\z)^\top\hat{s}_{\textrm{STEIN},k}(\z)\right] - \E\left[\hat{s}_{\textrm{SI},k}(\z)^\top\hat{s}_{\textrm{SI},k}(\z)\right] \\
   &=  \E\left\Vert\hat{s}_{\textrm{STEIN},k}(\z)\right\Vert^2_2 - \E\left\Vert\hat{s}_{\textrm{SI},k}(\z)\right\Vert^2_2.
\end{align}
Since for the Gaussian density kernel $k$
\begin{equation}
    \nabla_{\zp}k(\z, \zp) = k(\z, \zp)\cdot\frac{\zp - \z}{\sigma_k^2},
\end{equation} and for the Gaussian conditional likelihood $\qzgiveneps$
\begin{equation}
    \nabla_{\zp}\log \qzgiveneps(\zp\vert\eps^\prime) = \mathrm{diag}(\bm{\sigma}_{\eps^\prime})^{-2}\left(\zp - \bm{\mu}_{\eps^\prime}\right),
\end{equation} it holds that
    \begin{align}
    \begin{split}
    \Delta\V &= \frac{1}{n} \E_{\zp \sim \qz}\left[ k(\z, \zp)^2\cdot\frac{\Vert\zp - \z\Vert^2}{\sigma_k^4}\right] \\
    &\quad -\frac{1}{n}\E_{\eps^\prime, \etav^\prime \sim \pepsetav}\left[k(\z, \mathrm{diag}(\bm{\sigma}_{\eps^\prime}) \etav^\prime + \bm{\mu}_{\eps^\prime})^2 \cdot  \Vert \mathrm{diag}(\bm{\sigma}_{\eps^\prime})^{-1} \etav^\prime\Vert^2_2\right] .
        \end{split} \\
            \begin{split}
& = \frac{1}{n}\E_{\eps^\prime, \etav^\prime \sim \pepsetav}k(\z, \mathrm{diag}(\bm{\sigma}_{\eps^\prime}) \etav^\prime + \bm{\mu}_{\eps^\prime})^2 \\
    &\quad\cdot \left[\frac{\Vert  \mathrm{diag}(\bm{\sigma}_{\eps^\prime}) \etav^\prime + \bm{\mu}_{\eps^\prime} - \z\Vert^2_2 }{\sigma_k^4} - \Vert \mathrm{diag}(\bm{\sigma}_{\eps^\prime})^{-1} \etav^\prime\Vert^2_2\right].
       \end{split}
    \end{align}

\subsection{Sufficient condition for lower score gradient variance}
\label{sec:proof_low_var_suf}

Since 
\begin{equation}
     \Delta\V =\mathop{\E}_{\eps^\prime, \etav^\prime \sim \pepsetav}\underbrace{\frac{k(\z, \mathrm{diag}(\bm{\sigma}_{\eps^\prime}) \etav^\prime + \bm{\mu}_{\eps^\prime})^2}{n}}_{\geq 0} \cdot\underbrace{ \left[\frac{\Vert  \mathrm{diag}(\bm{\sigma}_{\eps^\prime}) \etav^\prime + \bm{\mu}_{\eps^\prime} - \z\Vert^2_2 }{\sigma_k^4} - \Vert \mathrm{diag}(\bm{\sigma}_{\eps^\prime})^{-1} \etav^\prime\Vert^2_2\right]}_{=: \beta} ,
\end{equation}
it follows from $\beta \geq 0$ a.s. that $\Delta\V \geq 0.$

From this, we get that \begin{align}
    \beta \geq 0 \text{ a.s.} &\iff \frac{\Vert  \mathrm{diag}(\bm{\sigma}_{\eps^\prime}) \etav^\prime + \bm{\mu}_{\eps^\prime} - \z\Vert^2_2 }{\Vert \mathrm{diag}(\bm{\sigma}_{\eps^\prime})^{-1} \etav^\prime\Vert^2_2} \geq  \sigma_k^4 \text{ a.s.}  \\
    & \Leftarrow \frac{\Vert  \mathrm{diag}(\bm{\sigma}_{\eps^\prime}) \etav^\prime\Vert^2_2 + 2  (\mathrm{diag}(\bm{\sigma}_{\eps^\prime}) \etav^\prime)^\top(\bm{\mu}_{\eps^\prime} - \z) + \Vert \bm{\mu}_{\eps^\prime} - \z\Vert^2_2}{\frac{1}{\min(\bm{\sigma}_{\eps^\prime})^2}\Vert\etav^\prime\Vert^2_2}  \geq  \sigma_k^4 \text{ a.s.} \\
        & \Leftarrow \frac{\min(\bm{\sigma}_{\eps^\prime})^2\Vert  \etav^\prime\Vert^2_2 - 2  \Vert(\mathrm{diag}(\bm{\sigma}_{\eps^\prime}) \etav^\prime)\Vert_2\Vert(\bm{\mu}_{\eps^\prime} - \z)\Vert_2 + \Vert \bm{\mu}_{\eps^\prime} - \z\Vert^2_2}{\frac{1}{\min(\bm{\sigma}_{\eps^\prime})^2}\Vert\etav^\prime\Vert^2_2}  \geq  \sigma_k^4 \text{ a.s.}\\
       &\Leftarrow \min(\bm{\sigma}_{\eps^\prime})^4 - 2\min(\bm{\sigma}_{\eps^\prime})^2\max(\bm{\sigma}_{\eps^\prime})\frac{\Vert  \bm{\mu}_{\eps^\prime} - \z\Vert_2}{\Vert\etav^\prime\Vert_2}  +  \min(\bm{\sigma}_{\eps^\prime})^2\frac{\Vert  \bm{\mu}_{\eps^\prime} - \z\Vert^2_2}{\Vert\etav^\prime\Vert^2_2}\geq \sigma_k^4 \quad \text{a.s.}
\end{align}

\subsection{Upper bound for the difference in score gradient variance}
\label{sec:proof_ub_diff_sgv}
 Under the assumptions that $\alpha = k(\z, \zp)^2$ and $\beta = \left[\frac{\Vert  \mathrm{diag}(\bm{\sigma}_{\eps^\prime}) \etav^\prime + \bm{\mu}_{\eps^\prime} - \z\Vert^2_2 }{\sigma_k^4} - \Vert \mathrm{diag}(\bm{\sigma}_{\eps^\prime})^{-1} \etav^\prime\Vert^2_2\right]$ are negatively correlated, it holds that
         \begin{align}
         \Delta\V &=  \E\left[\alpha\beta\right]\\
         &\leq \E\left[\alpha\right]\cdot\E\left[\beta\right]\\
    \begin{split}
    & = \frac{1}{n}\E_{\eps^\prime, \etav^\prime \sim \pepsetav}k(\z, \mathrm{diag}(\bm{\sigma}_{\eps^\prime}) \etav^\prime + \bm{\mu}_{\eps^\prime})^2\\
    &\cdot \mathop{\E}_{\eps^\prime, \etav^\prime \sim \pepsetav}\left[\frac{\Vert \bm{\sigma}_{\eps^\prime} \odot \etav^\prime\Vert^2_2 + 2(\bm{\sigma}_{\eps^\prime}\odot \etav^\prime)^\top(\bm{\mu}_{\eps^\prime} - \z) + \Vert\bm{\mu}_{\eps^\prime} - \z\Vert^2_2 }{\sigma_k^4} - \Vert \bm{\sigma}_{\eps^\prime}^{\odot-1} \odot\etav^\prime\Vert^2_2\right]
       \end{split} \\
    \begin{split}
     &\overset{\etav\sim\mathcal{N}(0, I)}{=} \frac{1}{n}\E_{\z\sim \qz}k(\z, \zp)^2\\
    &\cdot\underbrace{\left(\frac{\E_{\eps^\prime \sim \peps}\left[\Vert  \bm{\sigma}_{\eps^\prime}\Vert^2_2 + \Vert\bm{\mu}_{\eps^\prime} - \z\Vert^2_2\right] }{\sigma_k^4} - \E_{\eps^\prime \sim \peps}\left[\Vert \bm{\sigma}_{\eps^\prime}^{\odot-1}\Vert^2_2\right]\right)}_{=: \gamma} 
       \end{split}.\\
    &=  \frac{1}{n}\E_{\z\sim \qz}\left[\frac{1}{(2\pi\sigma_k^2)^{d_z}}\exp\left(-\frac{\left\Vert\z - \zp\right\Vert^2_2}{\sigma_k^2}\right)\right]\cdot \gamma\\
    &\approx \frac{\qz(\z)}{n(2\sqrt{\pi}\sigma_k)^{d_z}} \cdot \gamma.
    \end{align}

\subsection{Bounds of the optimal proposal distribution}
\label{sec:proof_rc1o}
For $\alpha(\z) = 1,$ we get the trivial upper bound solution \begin{equation}
    \tauepsgivenz^* = 1\cdot \peps(\eps) + 0 \cdot \tautildeepsgivenz^*(\epsgivenz) = \peps(\eps).
\end{equation}
For $\alpha(\z) = 0,$ assuming that $\tautildeepsgivenz$ is sufficiently flexible, we get that
\begin{align}
     &\tautildeepsgivenz^* \in \arg\min_{\tautildeepsgivenz} \,
\E_{\z, \eps \sim \qzeps} \left[
- \log \left(\tauepsgivenz(\epsgivenz)\qz(\z)\right)
\right] \\
\iff &  \tautildeepsgivenz^* \in \arg\min_{\tautildeepsgivenz} \,
\E_{\z, \eps \sim \qzeps} \left[
- \log \left(\tautildeepsgivenz(\epsgivenz)\qz(\z)\right)
\right] \\
\iff &  \tautildeepsgivenz^* \in \arg\min_{\tautildeepsgivenz} \,
\KL(\qzeps\Vert \tautildeepsgivenz \cdot \qz) \\
\iff & \tautildeepsgivenz^* \cdot \qz = \qzeps \\
\iff & \tautildeepsgivenz = \qepsgivenz \quad (=\tauepsgivenz).
\end{align}

%%%%%%%%%%%%%%%%%%%%%%%%%%%%%%%%%%%%%%%%%%%%%%%%%%%%%%%%%%%%
\newpage
\section{Further benchmarks} \label{app:furtherbenchmarks}

We evaluate all kernelized SIVI variants on the \textit{banana}, \textit{x-shaped}, and \textit{multimodal} benchmark distributions, originally proposed by \cite{cheng2024kernel}, with the corresponding target densities summarized in Table~\ref{tab:toy_densities}. We train each model for 500 epochs with 100 optimization steps per epoch and a batch size of 500. The architecture is a fully connected ReLU network with a latent dimension of 3, hidden layer size of 50, and output dimension of 2. Optimization is performed using the Adam optimizer with a learning rate of 0.001. A learning rate decay with factor 0.9 is applied every 1000 steps. For the \textit{x-shaped} and \textit{banana} targets, no annealing is used, while for the \textit{multimodal} target, annealing is enabled.

\begin{figure}[h]
\centering
    \includegraphics[width=\textwidth]{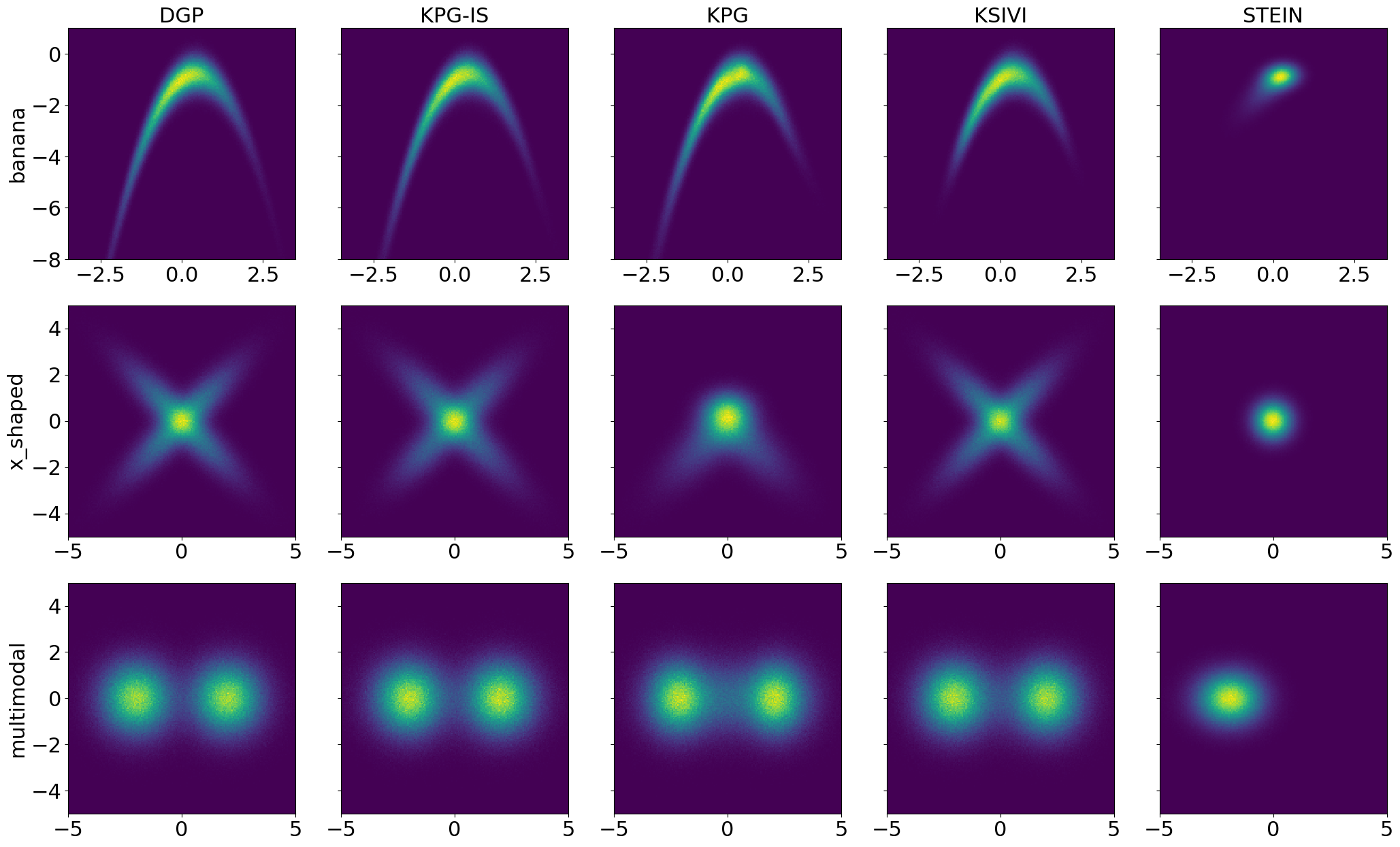}
\caption{Density contour plots for the DGP and each method from one representative repetition. 
Each plot visualizes the estimated density using 100{,}000 $\eps$-samples. 
The DGP serves as the reference distribution, while the others illustrate the approximation quality of each method.}
\label{fig:density_contours}
\end{figure}

The performance across three independent runs is presented in Table~\ref{tab:nll_results}, and representative contour plots of the final models are shown in Figure~\ref{fig:density_contours}. Among the tested methods, KPG-IS consistently outperforms all other kernel-based SIVI approaches. This is particularly evident on the banana benchmark, where the comparison with KSIVI highlights KPG-IS’s ability to more effectively capture narrow, high-density regions. Additionally, the comparison between Stein and KPG-IS clearly demonstrates the benefits of variance reduction through structured exploitation of the SIVI framework.

\begin{table}[h]
\caption{Definitions of the benchmark target distributions.}
\label{tab:toy_densities}
% \vskip -0.15in
% \begin{center}
\begin{small}
% \begin{sc}
\resizebox{0.99\textwidth}{!}{
\begin{tabular}{lcc}
\toprule
Benchmark & Distribution & Parameters \\
\midrule
Banana     & $z = (\nu_1, \nu_1^2 + \nu_2 + 1)^\top$, where $\nu \sim \mathcal{N}(0, \Sigma)$ & $\Sigma = \begin{bmatrix}1 & 0.9 \\ 0.9 & 1\end{bmatrix}$ \\
Multimodal & $z \sim \frac{1}{2} \mathcal{N}(\mu_1, I) + \frac{1}{2} \mathcal{N}(\mu_2, I)$ & $\mu_1 = (-2, 0)^\top$, $\mu_2 = (2, 0)^\top$ \\
X-shaped    & $z \sim \frac{1}{2} \mathcal{N}(0, \Sigma_1) + \frac{1}{2} \mathcal{N}(0, \Sigma_2)$ & $\Sigma_1 = \begin{bmatrix}2 & 1.8 \\ 1.8 & 2\end{bmatrix}$, $\Sigma_2 = \begin{bmatrix}2 & -1.8 \\ -1.8 & 2\end{bmatrix}$ \\
\bottomrule
\end{tabular}
}
% \end{sc}
\end{small}
% \end{center}
\vskip -0.1in
\end{table}

\begin{table}[h]
{\fontsize{8.5pt}{10pt}\selectfont
\centering
\caption{Benchmark comparison for problems in \cref{tab:toy_densities}. Each negative log likelihood (NLL) is computed on 100{,}000 samples from the data-generating process evaluated using model estimates from 100{,}000 $\eps$-samples.  
Reported are mean $\pm$ standard deviation over 3 runs. 
Best-performing methods (closest to DGP) are shown in \textbf{bold}.}
\label{tab:nll_results}
\resizebox{0.99\textwidth}{!}{
\begin{tabular}{l|c|cccc}
\toprule
Dataset & DGP (Ground Truth) & KPG-IS & KPG & KSIVI & STEIN \\
\midrule
banana     & $2.0024$  & $\mathbf{2.0913} \pm 0.0026$ & $2.1295 \pm 0.0341$ & $3.6227 \pm 0.2545$ & $8.1287 \pm 2.4536$ \\
x\_shaped  & $3.1219$  & $\mathbf{3.5973} \pm 0.8169$ & $4.0132 \pm 0.7670$ & $3.8713 \pm 1.2980$ & $7.9521 \pm 0.0980$ \\
multimodal & $3.4663$  & $\mathbf{3.4666} \pm 0.0001$ & $3.4703 \pm 0.0010$ & $3.4671 \pm 0.0001$ & $7.8133 \pm 0.5641$ \\
\bottomrule
\end{tabular}}
}
\end{table}
\normalsize

\section{Computational environment} \label{app:compsetup}
All experiments are performed on a Linux-based server A5000 server with 2 GPUs, 24GB VRAM, and Intel Xeon Gold 5315Y processor with 3.20 GHz.

\section{Experimental details}
\label{app:expdetails}
For all kernelized SIVI variants, we use the Gaussian kernel and a conditional Gaussian likelihood with diagonal covariance. The mean of the likelihood is given by the output of a fully connected neural network, which we refer to as the SIVI model, while the variance is represented by a learnable parameter vector.

\subsection{Bayesian logistic regression}
\label{app:blr}
We used a batch size of 100 for training, and the likelihood was evaluated using the full dataset without subsampling. The model is a fully connected ReLU network with a latent dimension of 10, hidden layer size of 100, and output dimension of 22. Both the parameters of the SIVI models and their variance vector of the conditional Gaussian likelihood were optimized using the Adam optimizer with a learning rate of 0.001. A learning rate decay with factor 0.9 was applied every 3000 steps. Training was performed for 200{,}000 iterations.

\begin{figure}[h]
    \centering
    \includegraphics[width=0.7\linewidth]{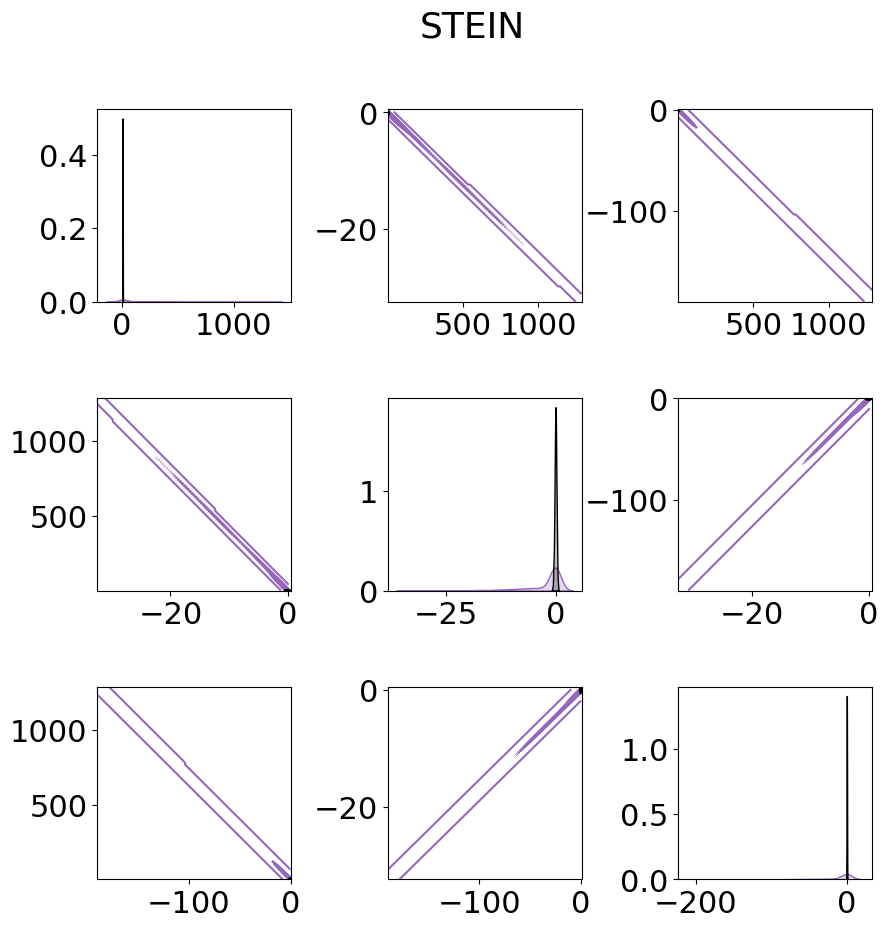}
    \caption{STEIN: Marginal and pairwise density estimates for $\bm{\beta}^{(1)}$, $\bm{\beta}^{(2)}$, and $\bm{\beta}^{(3)}$, with SGLD estimates shown in black for reference.}
    \label{fig:lr-coef-stein}
\end{figure}

We present representative marginal and pairwise distributions of the first three components for the Stein variant in Figure~\ref{fig:lr-coef-stein}, demonstrating that it fails to converge to the correct solution.

\newpage
\subsection{Conditional diffusion process}
We trained the model for 1000 epochs with 100 optimization steps per epoch, using a batch size of 128. The model is a fully connected ReLU network with a latent dimension of 100, hidden layer size of 128, and output dimension of 100. Both the parameters of the SIVI models and their variance vector of the conditional Gaussian likelihood were optimized using the Adam optimizer with a learning rate of 0.0002. A learning rate decay with factor 0.9 was applied every 10{,}000 steps. 

\begin{figure}
    \centering
    \includegraphics[width=0.6\linewidth]{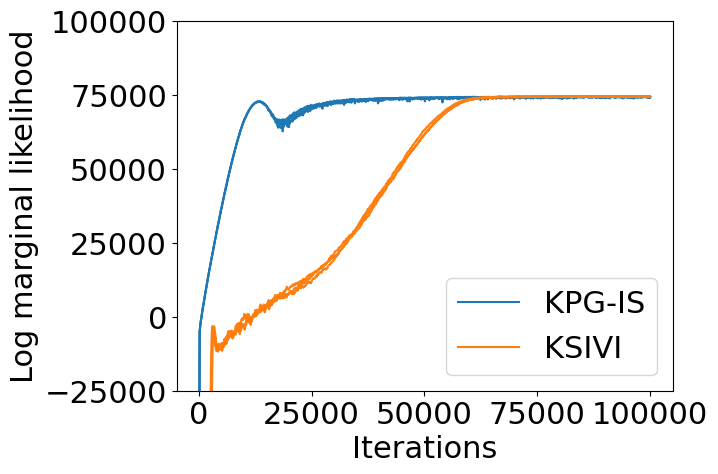}
    \caption{Convergence speed comparison between the current state-of-the-art method KSIVI and our proposal KPG-IS based on 3 repetitions.}
    \label{fig:langevin_perf_3}
\end{figure}

We repeat the 100-dimensional conditional diffusion process benchmark three times for the two best-performing models, KSIVI and KPG-IS, highlighting their stable training dynamics. As shown in Figure~\ref{fig:langevin_perf_3}, KPG-IS consistently demonstrates greater computational efficiency compared to KSIVI.

\label{app:cdp}

\section{Implementation details}
Note that each iteration of KPG requires $l$ target density evaluations, KSIVI requires $2l$, and KPG-IS requires $l^2$. When the cost of evaluating the target density becomes substantial, the quadratic cost of KPG-IS may become prohibitive. To address this, we employ the following adapted version of the KPG-IS algorithm:

In each iteration, we sample $\eps_j \sim \peps$ for $j = 1, \ldots, l$, and reuse these samples in place of $\eps_{i,j} \sim \peps$ wherever they would appear in Algorithm~\ref{alg:kpg_is}. By setting $\underline{\alpha}$ close to one, we achieve a substantial computational efficiency gain while still benefiting from the performance improvements afforded by partially informed sampling.

For further implementation details, we refer the reader to the accompanying code repository provided with this work.

\label{app:impdetails}
\end{document}